%% file: main.tex
\documentclass[sigconf,screen]{acmart}
\usepackage{hhline}
\usepackage{amsmath}

\usepackage{amssymb}
\usepackage{graphicx}
\usepackage{booktabs}
\AtBeginDocument{%
  }

\setcopyright{acmlicensed}
\copyrightyear{2026}
\acmYear{2026}
\acmDOI{XXXXXXX.XXXXXXX}
\acmConference[MM '26]{Proceedings of the 34th ACM International Conference on Multimedia}{November 10--14,
  2026}{Rio de Janeiro, Brazil}
\acmISBN{978-1-4503-XXXX-X/2018/06}
\renewcommand\footnotetextcopyrightpermission[1]{}

\newcommand{\first}[1]{\textbf{#1}}
\newcommand{\second}[1]{\underline{#1}}
\usepackage{booktabs}
\usepackage[table]{xcolor}
\definecolor{lightgray}{gray}{0.92}

\usepackage{multirow}
\acmSubmissionID{70}

\begin{document}

\title{GeoStereo: A Unified Stereo Geometry Estimation Framework for Disparity and Surface Normal}


\author{Qizhe Wei}
\authornote{These authors contributed equally to this work.}
\email{qizhewei@bit.edu.cn}
\affiliation{%
  \institution{Beijing Institute of Technology}
  \institution{BAAI}
  \city{Beijing}
  \country{China}
}

\author{Xianda Guo}
\authornotemark[1]
\email{xianda_guo@163.com}
\affiliation{%
  \institution{School of Computer Science, Wuhan University}
  \city{Wuhan}
  \country{China}
}

\author{Shaocong Xu}
\authornote{Project leader.}
\email{daniellesry@gmail.com}
\affiliation{%
  \institution{BAAI}
  \city{Beijing}
  \country{China}
}

\author{Hong Li}
\email{link0502@buaa.edu.cn}
\affiliation{%
  \institution{Beihang University}
  \institution{BAAI}
  \city{Beijing}
  \country{China}
}

\author{Runyi Yang}
\email{runyi.yang@insait.ai}
\affiliation{%
  \institution{INSAIT, Sofia University}
  \city{Sofia}
  \country{Bulgaria}
}

\author{Hao Zhao}
\authornote{Corresponding author.}
\email{zhaohao@air.tsinghua.edu.cn}
\affiliation{%
  \institution{BAAI}
  \institution{AIR, Tsinghua University}
  \city{Beijing}
  \country{China}
}

\renewcommand{\shortauthors}{Wei et al.}

\begin{teaserfigure}
  \centering
  \includegraphics[width=\textwidth]{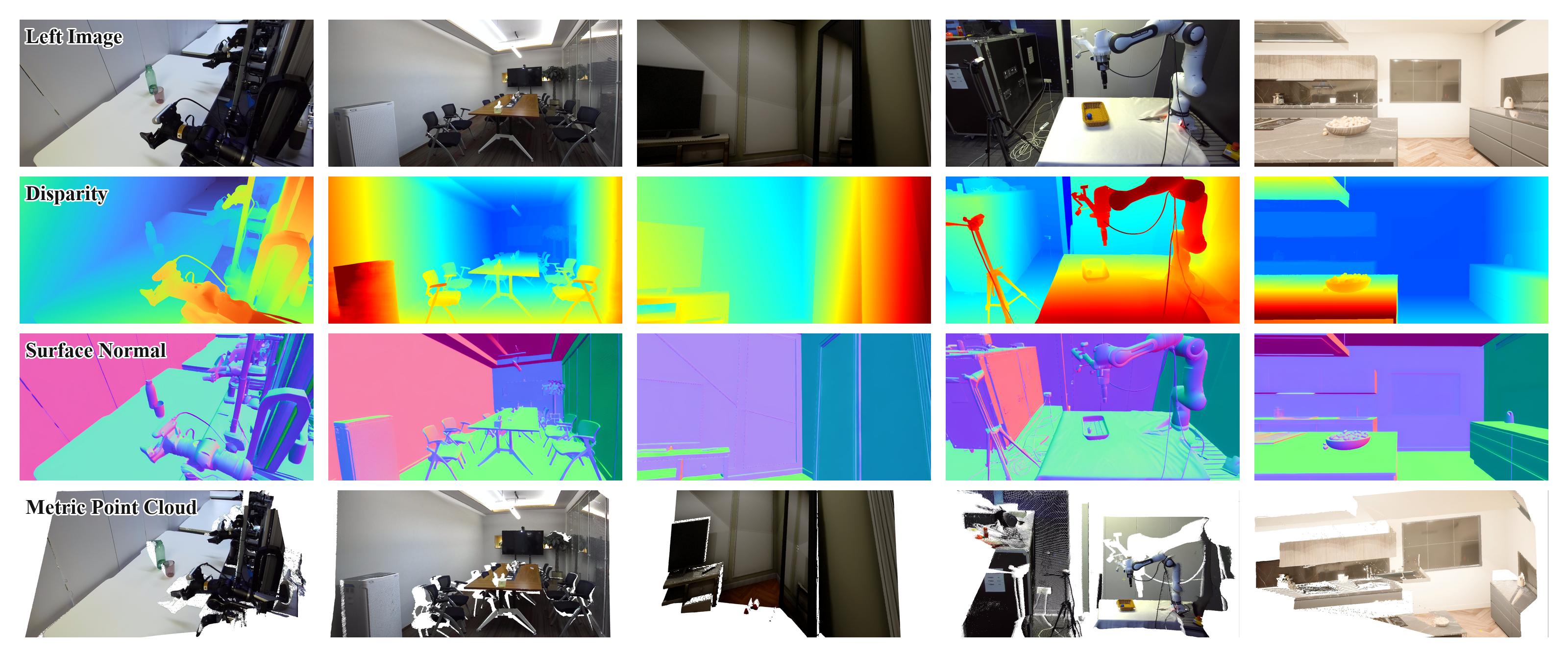}
  \caption{Zero-shot predictions on diverse indoor scenes, including robotics setups, office, reflective surfaces, and large open interiors. Given a stereo image pair, our method predicts disparity, surface normals, and metric point cloud reconstructions.}
  \label{fig:teaser}
\end{teaserfigure}

\input{MM26/0abstract}

\maketitle

\input{MM26/1introduction}

\input{MM26/2relatedwork}

\input{MM26/3method}

\input{MM26/4experiments}
\input{MM26/5conclusion}


\bibliographystyle{ACM-Reference-Format}
\bibliography{main}









\end{document}

%% file: MM26/0abstract.tex
\begin{abstract}
Stereo matching and surface normal estimation are fundamental tasks in 3D vision. However, existing feed-forward stereo methods still struggle to produce reliable predictions in challenging regions, mainly due to the lack of strong geometric priors. In this paper, we propose \textbf{GeoStereo}, a unified stereo geometry estimation framework that leverages powerful diffusion priors to jointly predict disparity and surface normals. Specifically, GeoStereo couples a feed-forward stereo matching pipeline with a diffusion-based normal estimation branch. To enable effective interaction between the two tasks, we introduce a disparity to normal initialization strategy and construct a warp to left-view condition for the diffusion process. This coupled design allows the diffusion branch to provide strong structural priors that enhance disparity estimation in ill-posed regions, while the feed-forward branch offers reliable geometric guidance for accurate normal prediction. Extensive experiments show that GeoStereo performs reliably in challenging scenarios, including low-light environments, highly reflective surfaces, and transparent objects. Under zero-shot settings, it achieves Rank-1 disparity estimation on multiple benchmarks, including KITTI and NYUv2, and delivers the best normal estimation accuracy on many real indoor benchmarks, such as iBims-1 and ScanNet. Project page: \url{https://qz-wei.github.io/GeoStereo.github.io/}
\end{abstract}

%% file: MM26/1introduction.tex
\section{Introduction}




Stereo matching is a fundamental task in 3D vision, aiming to estimate metric disparity from stereo images. As a core geometric representation, disparity describes cross-view correspondence and scene depth variation, and supports many downstream applications, including robotics~\cite{ibdslam,mvgformer,bip3d} and autonomous driving~\cite{afnet,dggt}. Surface normals characterize local surface orientation and also provide useful cues for illumination understanding~\cite{liu2025light}, benefiting applications such as relighting and geometric reconstruction~\cite{ye2025hi3dgen,kocsis2025intrinsix}. Together, they provide complementary descriptions of scene geometry and are both essential for robust 3D understanding.

\begin{figure}[t]
  \centering
  \includegraphics[width=\columnwidth]{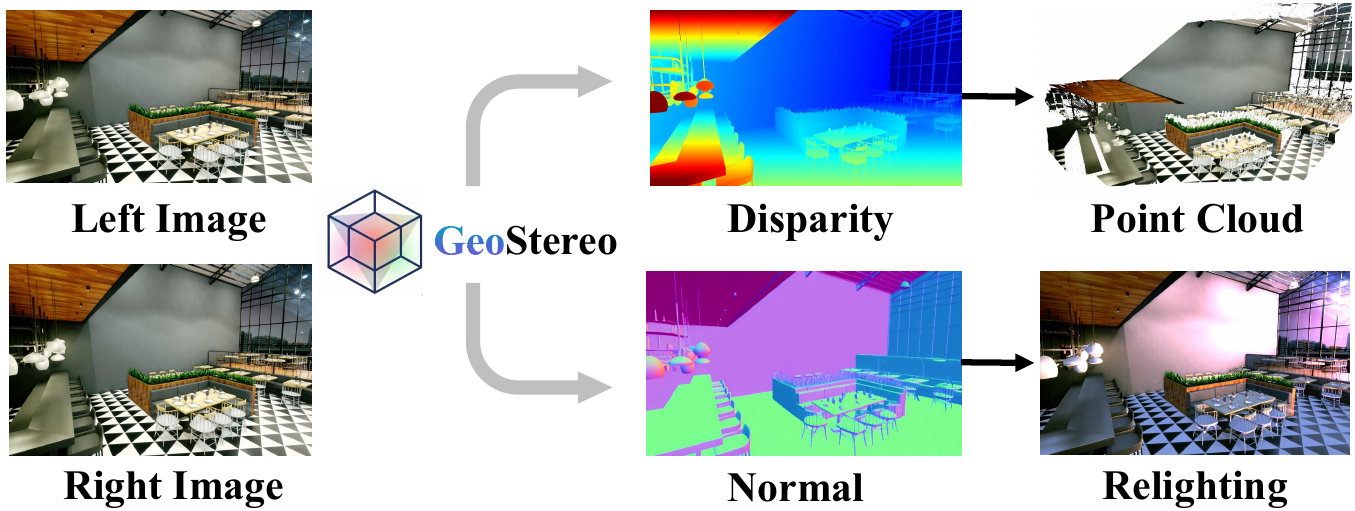}
\caption{GeoStereo jointly predicts disparity and surface normals from a stereo pair. The estimated geometry can be further used for metric point cloud and image relighting.}
\label{fig:point}
\end{figure}

Thanks to recent advances in large-scale datasets and foundation stereo models, such as FoundationStereo~\cite{wen2025stereo}, stereo matching has achieved impressive progress in both accuracy and generalization. However, existing methods still struggle to produce reliable predictions under challenging scenarios, such as extremely low-light environments, highly reflective surfaces, and transparent objects, as shown in Fig.~\ref{fig:dn}. These cases are common and important for downstream robotic applications, yet remain difficult for current feed-forward stereo pipelines. We attribute this phenomenon to the long-tail distribution of training data and the data-hungry nature of the feed-forward paradigm. To handle these challenging scenarios, in this work, we resort to powerful diffusion priors~\cite{stable_diffusion,ldm} and propose \textbf{GeoStereo}, as shown in Fig.~\ref{fig:point}, a new multi-task architecture for disparity and surface normal estimation.

The key idea is to explore the priors in diffusion models within a joint modeling paradigm that combines a feed-forward stereo matching network and a diffusion-based normal estimation network. Specifically, diffusion models have been demonstrated to contain powerful priors for dense prediction tasks, including monocular depth estimation~\cite{martingarcia2024diffusione2eft}, normal estimation~\cite{geowizard}, material estimation~\cite{ke2025marigold}, and image restoration~\cite{lin2025harnessing}. In particular, the normal map, which captures the local surface orientation, is used to relight the image. This property makes it a natural proxy task for activating the diffusion prior in low-light and weak-texture scenarios.

To enable joint optimization, GeoStereo incorporates two key
designs. First, the initial disparity is converted into a coarse
surface-normal estimate and injected into the diffusion branch as
a geometry-aware prior. Moreover, a right-view warped RGB condition
is constructed from the predicted disparity, providing spatially
aligned cross-view guidance. Since both pathways are differentiable
with respect to the initial disparity, gradients from the normal
objective can propagate back to the stereo branch. Consequently,
the diffusion branch introduces strong structural priors for
challenging regions, while stereo geometry provides reliable
guidance for normal prediction.


We demonstrate the effectiveness of GeoStereo across multiple zero-shot benchmarks for disparity and normal estimation. GeoStereo achieves state-of-the-art or highly competitive performance on both tasks, both quantitatively and qualitatively.


To summarize, our main contributions are as follows:
\begin{itemize}
    \item We propose \textbf{GeoStereo}, a unified framework that combines feed-forward stereo matching with diffusion-based normal estimation for disparity and surface normal prediction.
    
    \item We design an effective coupling mechanism between the two branches, enabling diffusion priors to enhance disparity estimation while allowing disparity to provide reliable geometric guidance for normal prediction.
    
    \item Extensive experiments demonstrate the effectiveness of our method, as well as its strong zero-shot generalization and robust performance in challenging scenarios on both tasks.
\end{itemize}

%% file: MM26/2relatedwork.tex
\begin{figure*}[t]
  \centering
  \includegraphics[width=\textwidth,height = 78mm]{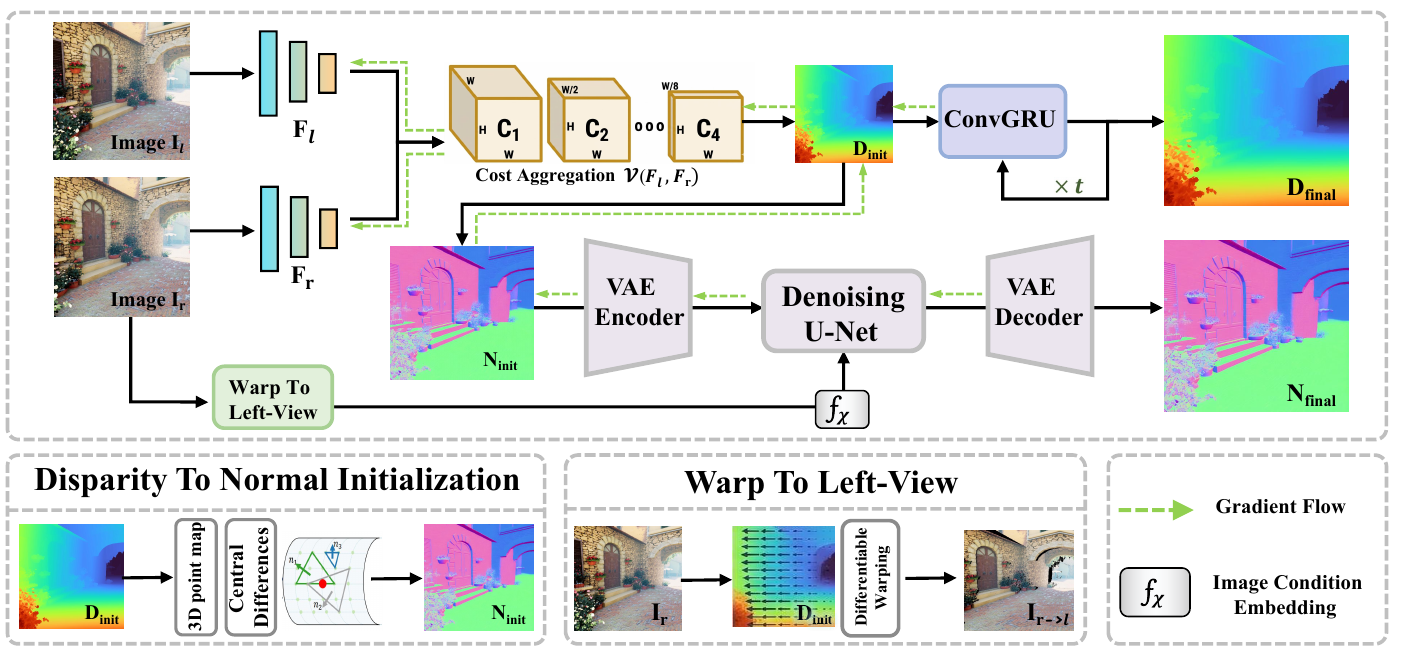}
\caption{
Overview of the proposed GeoStereo framework. A stereo pair is first
processed by a weight-shared stereo encoder, followed by cost
aggregation and ConvGRU-based refinement to predict the initial and
final disparities. The initial disparity is converted into a
geometry-aware normal prior and is also used to warp the right image
into the left view. The disparity to normal initialization and aligned
cross-view condition guide the diffusion-based U-Net, whose refined
latent is decoded to produce the final normal map.
}
\Description{Diagram showing the flow of stereo images through a shared encoder, aggregation, disparity prediction, ConvGRU refinement, and a diffusion-based U-Net to produce a normal map.}
\label{fig:pipe}
\end{figure*}
\section{Related Work}
\subsection{Learning-based Stereo Matching}
Deep stereo matching has substantially advanced the field by replacing hand-crafted features with learnable representations. Since the emergence of end-to-end stereo networks~\cite{gcnet,chang2018pyramid}, aggregation-based methods have become the dominant paradigm for stereo matching~\cite{sceneflow,gwcnet,ganet,leastereo,cfnet,acvnet,xu2023accurate,wang2024selective,wei2026aquastereo}. These methods generally follow a four-step pipeline of feature extraction, cost-volume construction, cost aggregation, and disparity regression, and have progressively improved both matching accuracy and computational efficiency. After that, iterative-based methods became a major research direction~\cite{raftstereo,crestereo,igev,igevpp,nmrf,guo2025lightstereo,zhang2023temporalstereo,guo2023openstereo}. Instead of relying on heavy cost aggregation, these methods utilize local cost lookup and iterative refinement to progressively update disparity, enabling more accurate and efficient high-resolution inference. More recently, many methods have begun to leverage vision foundation models~\cite{stereoanything,wen2025stereo,monster,s2m2,stereoanywhere,zerostereo,wang2026promptstereo,zhou2025all,zhang2025ctrl}, especially monocular depth foundation models~\cite{depth_anything_v2,guan2025bridgedepth}, to enhance stereo matching. Rather than only improving local matching modules, these approaches introduce stronger priors through robust feature extractors~\cite{wang2025moe,jiang2025defom}, monocular guidance, or transferable pretrained representations, leading to better generalization across domains.


\subsection{Surface Normal Estimation}
Surface normal estimation is a fundamental task in 3D scene understanding. Early learning-based methods mainly formulate it as a regression problem~\cite{eigen2015predicting,huang2019framenet,bae2021estimating,omnidata}, where convolutional or transformer-based networks directly predict normal maps from input images. Representative works range from early multi-task CNN architectures~\cite{eigen2015predicting,long2024adaptive,liu2024rip} to methods that learn local canonical frames~\cite{huang2019framenet,xu2025diffusion}, estimate aleatoric uncertainty for better boundary modeling~\cite{bae2021estimating,chen2022cerberus}, and exploit large-scale pretraining with stronger geometric supervision~\cite{omnidata,dsine,chen2026light}. With stronger architectures and larger-scale training, these regression-based methods have achieved promising performance, but they often struggle to preserve fine geometric details and generalize to challenging real-world scenes. More recently, diffusion models~\cite{ldm,diffusion_transferable} have opened a new direction for normal estimation by repurposing large pretrained image generators as geometric predictors. A common paradigm is to formulate normal estimation as an image-to-image translation task and fine-tune latent diffusion backbones with additional image-conditioning branches~\cite{stable_diffusion,controlnet}. Representative works~\cite{li2025light,genpercept,hu2024surface,wang2025fe2e}, such as GeoWizard~\cite{geowizard}, StableNormal~\cite{ye2024stablenormal} and Marigold~\cite{ke2025marigold} demonstrate that diffusion priors can produce sharp and detailed normal maps.

\subsection{Joint Disparity/Depth and Surface Normal Learning}

Joint learning of depth/disparity and surface normals has attracted increasing attention in recent years~\cite{hu2024metric3dv2,wang2025moge2}, since the two tasks describe complementary aspects of scene geometry. Prior works have shown that surface normals provide informative local priors for depth recovery, while depth offers useful structural cues for normal estimation~\cite{depthandsurface,Normal-distance_assisted_monoculardepth}. However, most existing methods are developed in monocular settings, where depth estimation often suffers from scale ambiguity and is commonly formulated in a relative-depth manner. In contrast, stereo disparity is directly grounded in binocular geometry and can be converted into metric depth and point clouds under known camera parameters, making joint disparity-normal modeling attractive for accurate 3D reconstruction. Nevertheless, such joint learning remains much less explored in stereo or multi-view settings. This limitation is especially evident in challenging regions, such as weakly textured areas, reflective surfaces, and object boundaries, where stereo correspondence is often ambiguous while normal cues can provide complementary guidance. Earlier attempts typically introduce normal-related constraints only in post-processing or optimization~\cite{irs,liu2022digging}, rather than within an end-to-end unified framework, leaving the complementary relationship between disparity and surface normals underexplored.

%% file: MM26/3method.tex
\begin{figure*}[t]
  \centering
  \includegraphics[width=\textwidth]{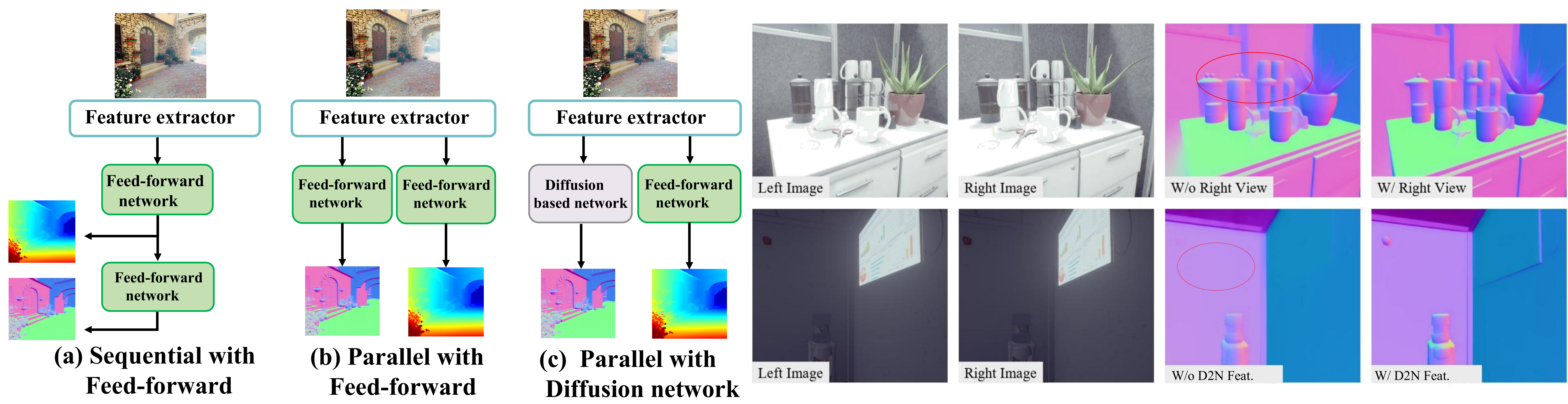}
\caption{Left: Different architectures for integrating disparity and normal estimation. Right: Effects of right-view guidance and disparity to normal initialization. ``W/o Right View'' uses only the left-view observation, while ``W/o D2N'' removes the disparity to normal initialization. The results demonstrate that aligned cross-view cues improve structural boundaries, while D2N provides a reliable geometric starting point for diffusion-based normal refinement.
}
    \label{fig:compare_arch}
\end{figure*}

\section{Method}
In this section, we introduce the problem formulation and the overall design of our method. As illustrated in Fig.~\ref{fig:pipe}, our framework jointly models disparity estimation and surface normal estimation within a unified stereo framework.

\subsection{Overview and Problem Formulation}

The key idea of our framework is to build a unified stereo geometry estimation model that jointly predicts disparity and surface normals through complementary modeling paradigms. In challenging stereo scenarios, such as weakly textured, reflective, or transparent regions, disparity estimation often becomes unreliable because correspondence is ambiguous and such cases are underrepresented in labeled stereo datasets. In contrast, diffusion-based normal estimation benefits from strong diffusion priors learned from large-scale image data, which provide richer structural cues for geometry reasoning. Motivated by this observation, we use the diffusion branch to introduce strong diffusion priors that complement disparity estimation, while the predicted disparity provides structured geometric guidance for normal generation. In this way, GeoStereo explicitly couples the two tasks, promoting the integration of their complementary priors and enabling more robust geometry estimation.

As illustrated in Fig.~\ref{fig:pipe}, our framework consists of two main components. First, the stereo matching branch extracts correspondence-aware features from the rectified stereo pair, constructs a cost volume, and predicts an initial disparity together with a refined final disparity. Second, the initial disparity is converted into a coarse surface-normal estimate and is also used to differentiably warp the right image into the left view. These two representations provide complementary geometric and cross-view guidance for normal estimation.

\subsection{Preliminaries on Diffusion Model}

Diffusion Probabilistic Models aim to model a data distribution $p(x)$ by sequentially transforming a Gaussian distribution via a so-called backward diffusion process. This backward process is uniquely determined by a predefined forward diffusion process. 

Taking DDPM~\cite{ho2020denoising} as a classic example, the forward diffusion process gradually adds noise to the initial data $x_0 \sim p(x)$. The noisy data at time step $t \in \{1, \dots, T\}$ can be directly sampled via:
\begin{equation}
q(x_t) = \sqrt{\bar{\alpha}_t} x_0 + \sqrt{1 - \bar{\alpha}_t}\epsilon, \quad \epsilon \sim \mathcal{N}(0, \mathbf{I}).
\end{equation}
where $T$ denotes the total number of time steps, and $\bar{\alpha}_t$ is the cumulative noise schedule controlling how fast the data distribution is transformed into a standard Gaussian distribution. As a result, the backward diffusion process in DDPM is formulated to reverse this corruption:
\begin{equation}
x_{t-1} = \frac{1}{\sqrt{\alpha_t}} x_t - \frac{1 - \alpha_t}{\sqrt{\alpha_t(1 - \bar{\alpha}_t)}} \mu_\theta^\epsilon(x_t, t) + \sigma_t \epsilon.
\end{equation}
where $\mu_\theta^\epsilon$ is a neural network trained to predict the injected noise, and $\sigma_t$ controls the variance of the reverse step.

\noindent\textbf{Reparameterization.} 
It is often convenient to reparameterize diffusion models based on their prediction targets. By default, the network is trained to predict the injected noise (referred to as $\epsilon$-reparameterization), optimized by the following loss function:
\begin{equation}
L_\theta = \mathbb{E}_{\epsilon, c, t} \|\epsilon - \mu_\theta^\epsilon(x_t, c, t)\|^2.
\end{equation}
However, for certain image-to-image translation tasks, it is more straightforward to reparameterize the model to directly predict the one-step denoised output (known as $x_0$-reparameterization). Under this setting, the loss function is equivalently formulated as a denoising autoencoder loss:
\begin{equation}
L_\theta = \mathbb{E}_{x_0, c, t} \|x_0 - \mu_\theta^{x_0}(x_t, c, t)\|^2.
\end{equation}
where $c$ denotes the task-specific conditional inputs.

\noindent\textbf{Diffusion Samplers.} 
When the number of time steps $T$ is large enough, both the forward and backward diffusion processes can be viewed as approximations of continuous stochastic differential equations (SDEs). Therefore, it is possible to sample from a trained DDPM model using alternative SDE solvers or samplers for better efficiency, albeit at a slight cost of precision. For example, DDIM~\cite{song2020denoising} generates samples utilizing a non-Markovian process:
\begin{equation}
x_{t-1} = \sqrt{\bar{\alpha}_{t-1}} \cdot \left( \frac{x_t - \sqrt{1 - \bar{\alpha}_t} \cdot \mu_\theta^\epsilon(x_t, c, t)}{\sqrt{\bar{\alpha}_t}} \right)+ \text{direction}(x_t) + \tau \epsilon.
\end{equation}
where $\tau$ is a scalar controlling the amount of stochastic noise injected during the generation process. Notably, if $\tau$ is set to 0, DDIM becomes a completely deterministic sampler, rendering the sampling trajectory independent of any random noise.

\begin{figure*}[t]
  \centering
  \includegraphics[width=\textwidth]{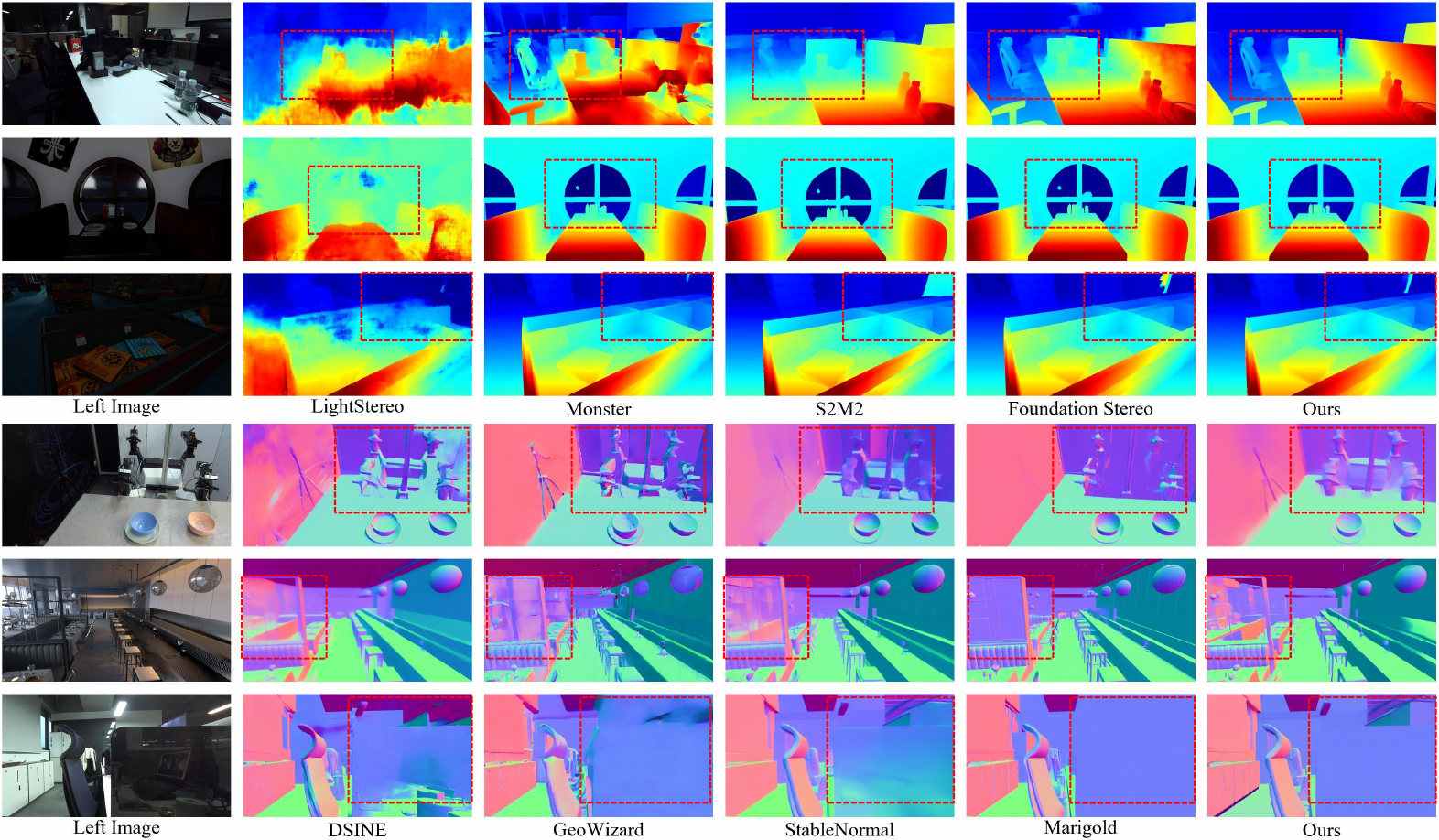}
\caption{Qualitative comparison of zero-shot inference on diverse scenes. All methods are trained on a mixture of public datasets. The examples include challenging cases such as low-light conditions, reflections, and transparent objects. The top three rows show disparity results, while the bottom row shows surface normal predictions.}
\label{fig:dn}
\end{figure*}

\subsection{Stereo Matching and Disparity to Normal Initialization}
\label{sec:disp2normal}

\noindent\textbf{Stereo Matching.}
Given a rectified stereo pair $(I_l, I_r)$, we first extract left and right features using a feature extractor initialized from strong monocular priors, e.g., DepthAnythingV2~\cite{depth_anything_v2}, and build a hybrid cost volume following FoundationStereo~\cite{wen2025stereo}:
\begin{equation}
F_l = \mathcal{E}(I_l), \qquad F_r = \mathcal{E}(I_r), \qquad \mathcal{C} = \mathcal{V}(F_l, F_r).
\end{equation}
Here, $\mathcal{E}(\cdot)$ denotes the feature extractor, $\mathcal{V}(\cdot)$ denotes the hybrid cost-volume construction and aggregation operator, and $\mathcal{C}$ denotes the resulting cost volume. Based on cost volume $\mathcal{C}$, we first estimate an initial disparity by soft argmin, and then refine it with a ConvGRU-based update module to obtain the final disparity:
\begin{equation}
D_{init} = \mathrm{SoftArgmin}(\mathcal{C}), \qquad D_{final} = \mathcal{U}_{\mathrm{gru}}(D_{init}, F_l).
\end{equation}
Here, $D_{init}$ denotes the initial disparity prediction obtained from cost-volume matching, while $D_{final}$ is the refined disparity produced by the ConvGRU update module $\mathcal{U}_{\mathrm{gru}}(\cdot)$.

\noindent\textbf{Disparity Loss.}
For disparity estimation, we supervise both the initial disparity and the final disparity using the $\ell_1$ loss:
\begin{equation}
\mathcal{L}_{\mathrm{disp}}
=
\lambda_{\mathrm{init}}
\left\|
D_{\mathrm{init}}-D_{\mathrm{gt}}
\right\|_1
+
\left\|
D_{\mathrm{final}}-D_{\mathrm{gt}}
\right\|_1.
\end{equation}
Here, $D_{\mathrm{gt}}$ denotes the ground-truth disparity, and $\lambda_{\mathrm{init}}$ balances the supervision on the initial disparity prediction.

\noindent\textbf{Disparity to Normal Initialization.}
To explicitly couple disparity and surface normal estimation and
reduce the ambiguity of diffusion-based normal prediction, we convert
$\mathbf{D}_{\mathrm{init}}$ into a coarse normal prior in a
geometry-motivated manner. The key observation is that disparity and surface normals are intrinsically related through local surface variation: a disparity field induces a 3D surface, whose normal can be estimated from local tangential directions. Specifically, we first transform disparity into a relative depth-like representation:
\begin{equation}
\bar{D}(u,v)=\mathrm{clip}\!\left(\frac{D_{init}(u,v)}{d_{\max}},\,\epsilon,\,1\right),
\quad
Z(u,v)=-\log \bar{D}(u,v).
\end{equation}
where $d_{\max}$ denotes the maximum disparity and $\epsilon$ is a small constant for numerical stability. In practice, we further smooth $Z$ with a local average filter before normal computation to suppress high-frequency noise. Using normalized image coordinates $(x(u),y(v))\in[-1,1]^2$, we define a pseudo-3D point map as:
\begin{equation}
\mathbf{P}(u,v)=
\begin{bmatrix}
x(u)\,Z(u,v)\\
y(v)\,Z(u,v)\\
Z(u,v)
\end{bmatrix}.
\end{equation}
We then estimate local tangential directions by central differences along the horizontal and vertical directions, and compute the initial normal as:
\begin{equation}
\tilde{\mathbf{N}}_{init}(u,v)=\mathbf{t}_x(u,v)\times \mathbf{t}_y(u,v),
\hspace{0.5em}
\mathbf{N}_{init}(u,v)=\frac{\tilde{\mathbf{N}}_{init}(u,v)}
{\|\tilde{\mathbf{N}}_{init}(u,v)\|_2}.
\end{equation}
Finally, we align the normal orientation to a consistent hemisphere and axis convention. In this way, $\mathbf{N}_{\mathrm{init}}$ provides a structured geometry-aware prior for subsequent diffusion-based normal estimation, while naturally maintaining geometric consistency.

\subsection{Diffusion-based Normal Estimator}
\label{sec:normal_estimate}

\noindent\textbf{Warp to Left-View.}
To further exploit binocular information, we introduce an aligned
right-view condition for diffusion-based normal refinement. Using only
the left image may be insufficient in challenging regions where local
textures are weak, reflective, transparent, or partially occluded.
The right image provides complementary observations, but directly
feeding it into the normal estimator introduces additional difficulty
because its content is not spatially aligned with the left reference
view.

We therefore warp the right image into the left-view coordinate system
using the initial disparity:
\begin{equation}
    \mathbf{I}_{r\rightarrow l}(u,v)
    =
    \mathcal{W}
    \left(
        \mathbf{I}_{r},
        \mathbf{D}_{\mathrm{init}}
    \right)(u,v)
    =
    \mathbf{I}_{r}
    \left(
        u-\mathbf{D}_{\mathrm{init}}(u,v),v
    \right),
\end{equation}
where $\mathcal{W}(\cdot)$ denotes differentiable backward warping
implemented with bilinear sampling. The aligned image
$\mathbf{I}_{r\rightarrow l}$ preserves complementary cross-view
appearance information while remaining spatially consistent with the
left-view normal prediction.

\noindent\textbf{Diffusion-based Normal Refinement.}
The disparity to normal initialization $\mathbf{N}_{\mathrm{geo}}$ provides a
structured geometry-aware prior for the diffusion-based normal
estimator. It is injected into the denoising U-Net as geometric
guidance, allowing the diffusion model to progressively correct local
artifacts and recover fine-grained surface structures. Compared with
relying only on the noisy latent, this prior reduces the ambiguity of
normal estimation.

Meanwhile, the warped right-view image
$\mathbf{I}_{r\rightarrow l}$ is processed by the conditional branch
$f_{\chi}(\cdot)$, providing spatially aligned cross-view appearance
guidance. Both the disparity to normal initialization and the
warp to left-view operation are differentiable with respect to
$\mathbf{D}_{\mathrm{init}}$. Therefore, gradients from the normal
objective can propagate back to the stereo branch through these two
pathways. This allows diffusion priors to regularize disparity
estimation, while stereo geometry provides reliable guidance for
normal refinement.
\noindent\textbf{Surface Normal Loss.}
Following I2VGen-XL~\cite{zhang2023i2vgen}, we adopt the
$x_0$-reparameterization for the denoising U-Net. During training,
the ground-truth normal map $\mathbf{N}_{\mathrm{gt}}$ is encoded into
the clean latent $x_0$, which is perturbed according to the predefined
diffusion noise schedule to obtain $x_t$. The normal estimation
objective is defined as
\begin{equation}
\mathcal{L}_{\mathrm{norm}}
=
\mathbb{E}_{x_0,t,\epsilon}
\left[
\left\|
x_0-
\mu_{\zeta}^{x_0}
\bigl(
x_t,\,
t,\,
\mathbf{N}_{\mathrm{init}},\,
f_{\chi}(\mathbf{I}_{r\rightarrow l})
\bigr)
\right\|_2^2
\right].
\end{equation}

After iterative denoising, the refined latent is decoded by the VAE
decoder to obtain the final normal prediction.

\noindent\textbf{Consistency Training Strategy.}
Our framework is further unified at the optimization level through a consistency training strategy. In practice, we adopt a two-stage training scheme. In the first stage, the disparity branch and the normal branch are optimized separately for better efficiency and lower memory consumption, so as to obtain stable initial predictions for each task. In the second stage, the whole framework is jointly optimized to enforce consistency between disparity and normal estimation. Based on the disparity loss defined in Sec.~\ref{sec:disp2normal} and the normal estimation objective defined in Sec.~\ref{sec:normal_estimate}, the overall training objective is written as: 
\begin{equation}
\mathcal{L}
=
\mathcal{L}_{\mathrm{disp}}
+
\lambda_{\mathrm{norm}} \mathcal{L}_{\mathrm{norm}}.
\end{equation}
where $\lambda_{\mathrm{norm}}$ balances the contribution of the normal estimation objective. This consistency training enables the two branches to be progressively aligned during joint optimization, leading to more stable learning and more consistent predictions for both tasks.

%% file: MM26/4experiments.tex
\section{Experiments}

In this section, we compare our model with other SOTAs (i.e., FoundationStereo~\cite{wen2025stereo}, S2M2~\cite{s2m2}, Monster~\cite{monster}, 
DSINE~\cite{dsine}, StableNormal~\cite{ye2024stablenormal}, GeoWizard~\cite{geowizard} and Marigold~\cite{ke2025marigold}) in various real-world and synthetic datasets. In addition, an ablation study is conducted to demonstrate
the effectiveness of different components, i.e., LightStereo~\cite{guo2025lightstereo} and right-image.


\subsection{Implementation Details}
Our framework is implemented in PyTorch. We initialize the stereo module with pretrained FoundationStereo~\cite{wen2025stereo} weights and fine-tune it using AdamW~\cite{adam}, with an initial learning rate of $2\times10^{-5}$ that is decayed by a factor of 0.1 after 80\% of the total training schedule. Training is conducted with a total batch size of 16 on 4 NVIDIA H20 GPUs. The normal refinement module is built upon Stable Diffusion v2.1~\cite{stable_diffusion} and optimized with AdamW~\cite{adam} using a fixed learning rate of $1\times10^{-5}$. It is trained with a total batch size of 32 on 4 NVIDIA H20 GPUs.

\subsection{Datasets and Metrics}

\noindent\textbf{Datasets.}
Following FoundationStereo, we train the disparity component on a large mixed dataset consisting of FSD~\cite{wen2025stereo}, SceneFlow~\cite{sceneflow}, Sintel~\cite{singtel}, CREStereo~\cite{crestereo}, FallingThings~\cite{tremblay2018falling}, StereoCarlaV2~\cite{guo2025stereocarla}, IRS~\cite{irs}, 3D Ken Burns~\cite{3dken}, Structured3D~\cite{Structured3D}, InteriorVerse~\cite{inter}, DIODE~\cite{diode_dataset}, Virtual KITTI 2~\cite{cabon2020virtual}, as well as our self-collected dataset, totaling about 1.5M image-disparity pairs.

Following Marigold~\cite{ke2025marigold}, the normal component is trained on a large-scale dataset of high-resolution images with ground-truth surface normals rendered from synthetic scenes. The training data includes HyperSim~\cite{hypersim}, 3D Ken Burns~\cite{3dken}, Structured3D~\cite{Structured3D}, IRS~\cite{irs}, InteriorVerse~\cite{inter}, MatrixCity (small)~\cite{li2023matrixcity}, and our newly rendered dataset containing 60,000 images from 200 scenes. Most samples are photorealistically rendered with Blender, resulting in over 0.6M image-normal pairs in total.

\noindent\textbf{Metrics.}
We evaluate disparity prediction using End-Point Error (EPE) and the D1 outlier rate. EPE measures the average absolute disparity error over all valid pixels, while D1 denotes the percentage of pixels whose disparity error exceeds $\max(3\text{ px},\,0.05D)$. For surface normal estimation, we compute the angular error between the predicted and ground-truth normal maps. We report the mean and median angular errors, together with the percentage of pixels whose angular error is below $11.25^\circ$, $22.5^\circ$, and $30^\circ$.

\begin{table*}[t]
\centering
\caption{Quantitative evaluation. We compare our method with LightStereo~\cite{guo2025lightstereo}, Monster~\cite{monster}, S2M2~\cite{s2m2} and FoundationStereo~\cite{wen2025stereo}. Zero-shot generalization results are reported on multiple public stereo benchmarks.}
\label{tab:disp_zero_shot}

\setlength{\tabcolsep}{5pt}

\begin{tabular}{l|c||cc|cc|cc|cc|cc}
\toprule

\multirow{2}{*}{\centering Method} & \multirow{2}{*}{\centering Publication}
& \multicolumn{2}{c|}{Hypersim~\cite{hypersim}} 
& \multicolumn{2}{c|}{DREDS~\cite{dai2022dreds}} 
& \multicolumn{2}{c|}{KITTI~\cite{geiger2012kitti}}
& \multicolumn{2}{c|}{ScanNet~\cite{scannet}}
& \multicolumn{2}{c}{NYUv2~\cite{nyu}} \\
\hhline{~~||--|--|--|--|--}
\hhline{~~||--|--|--|--|--}
& 
& EPE$\downarrow$ & D1$\downarrow$ 
& EPE$\downarrow$ & D1$\downarrow$ 
& EPE$\downarrow$ & D1$\downarrow$
& EPE$\downarrow$ & D1$\downarrow$
& EPE$\downarrow$ & D1$\downarrow$ \\

\specialrule{0.5pt}{0pt}{0pt}
\specialrule{0.5pt}{1pt}{0pt}

LightStereo~\cite{guo2025lightstereo}
& ICRA'25
& 2.4881 & 6.07\%
& 0.4079 & 1.01\%
& 0.9244 & 3.53\%
& 1.7835 & 7.54\%
& 3.4842 & 17.57\% \\

Monster~\cite{monster}
& CVPR'25
& 2.5305 & 6.09\%
& 0.3155 & 0.70\%
& 1.2977 & 6.34\%
& 1.8869 & 7.71\%
& 3.0810 & 14.45\% \\

S2M2~\cite{s2m2}
& ICCV'25
& 0.8426 & 2.79\%
& 0.1025 & 0.27\%
& 1.3425 & 7.26\%
& \first{1.0559} & \first{3.82\%}
& 2.6500 & 8.08\% \\

Foundation Stereo~\cite{wen2025stereo}
& CVPR'25
& \second{0.4618} & \second{1.93\%}
& \second{0.0752} & \second{0.21\%}
& \second{0.8679} & \second{2.96\%}
& 1.0784 & 4.32\%
& \second{1.3922} & \second{6.54\%} \\

\rowcolor{cyan!10}
Ours
& -
& \first{0.3979} & \first{1.64\%}
& \first{0.0739} & \first{0.19\%}
& \first{0.8642} & \first{2.88\%}
& \second{1.0655} & \second{4.10\%}
& \first{1.3426} & \first{6.48\%} \\

\bottomrule
\end{tabular}
\end{table*}

\begin{table*}[t]
\centering
\small
\caption{Quantitative evaluation. We compare our method with DSINE~\cite{dsine}, GeoWizard~\cite{geowizard}, StableNormal~\cite{ye2024stablenormal} and Marigold~\cite{ke2025marigold}. Zero-shot generalization results are reported on four indoor normal benchmarks.}
\label{tab:normal_zero_shot}

\setlength{\tabcolsep}{4pt}
\renewcommand{\arraystretch}{1.12}
\arrayrulecolor{black}
\setlength{\arrayrulewidth}{0.8pt}

\begin{tabular}{l|c||ccccc|ccccc}
\toprule

\multirow{2}{*}{\centering Method} & \multirow{2}{*}{\centering Publication}
& \multicolumn{5}{c|}{iBims-1~\cite{iBims-1}}
& \multicolumn{5}{c}{NYUv2~\cite{nyu}} \\
\hhline{~~||-----|-----}
&
& Mean$\downarrow$ & Median$\downarrow$ & $<11.25^\circ\uparrow$ & $<22.5^\circ\uparrow$ & $<30^\circ\uparrow$
& Mean$\downarrow$ & Median$\downarrow$ & $<11.25^\circ\uparrow$ & $<22.5^\circ\uparrow$ & $<30^\circ\uparrow$ \\

\specialrule{0.5pt}{0pt}{0pt}
\specialrule{0.5pt}{1pt}{0pt}

DSINE~\cite{dsine}
& CVPR'24
& 17.36$^\circ$ & \second{6.97$^\circ$} & 66.81\% & 78.80\% & 82.23\%
& 16.18$^\circ$ & 8.92$^\circ$ & 59.64\% & 77.50\% & 83.35\% \\

GeoWizard~\cite{geowizard}
& ECCV'24
& 19.31$^\circ$ & 9.65$^\circ$ & 62.37\% & 77.12\% & 82.53\%
& 19.00$^\circ$ & 10.87$^\circ$ & 50.04\% & 72.95\% & 80.12\% \\

StableNormal~\cite{ye2024stablenormal}
& SIGGRAPH'24
& 18.91$^\circ$ & 8.62$^\circ$ & 62.21\% & 76.82\% & 81.26\%
& 20.63$^\circ$ & 12.15$^\circ$ & 49.75\% & 71.31\% & 78.21\% \\

Marigold~\cite{ke2025marigold}
& TPAMI'25
& \first{16.26$^\circ$} & \first{6.70$^\circ$} & \second{68.50\%} & \second{79.81\%} & \second{83.10\%}
& \second{15.58$^\circ$} & \second{8.54$^\circ$} & \second{61.63\%} & \second{78.45\%} & \second{84.17\%} \\

\rowcolor{cyan!10}
Ours
& -
& \second{17.00$^\circ$} & 7.02$^\circ$ & \first{68.92\%} & \first{80.14\%} & \first{83.58\%}
& \first{15.14$^\circ$} & \first{8.25$^\circ$} & \first{62.37\%} & \first{78.85\%} & \first{84.80\%} \\

\specialrule{0.5pt}{0pt}{0pt}

\multirow{1}{*}{\centering Method} & \multirow{1}{*}{\centering Publication}
& \multicolumn{5}{c|}{DIODE~\cite{diode_dataset}}
& \multicolumn{5}{c}{ScanNet~\cite{scannet}} \\

\specialrule{0.5pt}{0pt}{0pt}
\specialrule{0.5pt}{1pt}{0pt}

DSINE~\cite{dsine}
& CVPR'24
& 18.91$^\circ$ & 13.61$^\circ$ & 40.82\% & 76.68\% & 84.87\%
& 16.29$^\circ$ & 9.42$^\circ$ & 60.79\% & 78.40\% & 84.06\% \\

GeoWizard~\cite{geowizard}
& ECCV'24
& 22.74$^\circ$ & 15.82$^\circ$ & 37.10\% & 74.42\% & 80.85\%
& 17.64$^\circ$ & 9.82$^\circ$ & 54.68\% & 74.82\% & 81.20\% \\

StableNormal~\cite{ye2024stablenormal}
& SIGGRAPH'24
& 17.87$^\circ$ & \second{12.40$^\circ$} & \second{50.02\%} & \first{81.99\%} & \first{88.09\%}
& 19.31$^\circ$ & 14.47$^\circ$ & 47.70\% & 77.02\% & 84.42\% \\

Marigold~\cite{ke2025marigold}
& TPAMI'25
& \second{17.67$^\circ$} & 12.88$^\circ$ & 49.20\% & 80.95\% & 87.06\%
& \second{15.10$^\circ$} & \second{8.94$^\circ$} & \second{61.09\%} & \second{79.17\%} & \second{84.61\%} \\

\rowcolor{cyan!10}
Ours
& -
& \first{17.02$^\circ$} & \first{12.31$^\circ$} & \first{50.50\%} & \second{81.09\%} & \second{87.76\%}
& \first{14.21$^\circ$} & \first{8.60$^\circ$} & \first{62.55\%} & \first{79.70\%} & \first{85.31\%} \\

\bottomrule
\end{tabular}
\end{table*}

\subsection{Comparison to the state-of-the-art} 

\noindent\textbf{Zero-Shot Generalization Comparison.} Table~\ref{tab:disp_zero_shot} and Table~\ref{tab:normal_zero_shot} summarize the zero-shot comparison on disparity and surface normal estimation. Without any target-domain fine-tuning, our unified framework achieves state-of-the-art or highly competitive performance on most benchmarks. Since these datasets cover diverse synthetic and real-world scenarios with large variations in scene layout, appearance, materials, and imaging conditions, the results demonstrate the strong generalization ability of our method across domains. More importantly, as shown in Fig.~\ref{fig:dn}, the gains are particularly evident in challenging cases, showing that the proposed unified design is effective not only on standard benchmarks, but also in ill-posed regions that demand robust geometry estimation.

\noindent\textbf{Disparity Estimation.} On disparity estimation, we compare our method with representative stereo matching baselines, including LightStereo~\cite{guo2025lightstereo}, Monster~\cite{monster}, S2M2~\cite{s2m2}, and FoundationStereo~\cite{wen2025stereo}. These methods cover both efficient feed-forward architectures and recent foundation stereo models, providing a strong benchmark for evaluating the generalization ability of our unified framework. We conduct zero-shot evaluation on diverse datasets, including Hypersim~\cite{hypersim} and DREDS~\cite{dai2022dreds}, which are synthetic benchmarks with complex illumination and material properties, KITTI~\cite{geiger2012kitti}, which focuses on real-world outdoor driving scenes, and ScanNet~\cite{scannet} and NYUv2~\cite{nyu}, which contain real indoor environments with cluttered layouts and challenging depth discontinuities. As shown in Table~\ref{tab:disp_zero_shot}, our method achieves state-of-the-art or highly competitive zero-shot performance on most benchmarks. These results indicate that the proposed unified framework preserves the strong generalization ability of foundation stereo models, while further benefiting from the structural priors introduced by the diffusion branch. This advantage is especially important in ill-posed regions, where conventional feed-forward stereo matching is often unreliable. As illustrated in Fig.~\ref{fig:dn}, our method produces more reliable disparity predictions in challenging regions such as weakly textured areas, reflective surfaces, transparent objects, and object boundaries.


\noindent\textbf{Surface Normal Estimation.} The zero-shot results on surface normal estimation also demonstrate clear advantages over previous state-of-the-art methods. We compare our method with representative normal estimation approaches, including DSINE~\cite{dsine}, a strong feed-forward regression-based model, as well as recent diffusion-based methods such as GeoWizard~\cite{geowizard}, StableNormal~\cite{ye2024stablenormal}, and Marigold~\cite{ke2025marigold}. These methods represent the two dominant paradigms for normal prediction, namely deterministic regression and generative diffusion. We evaluate all methods under zero-shot settings on challenging indoor benchmarks, including iBims-1~\cite{iBims-1}, NYUv2~\cite{nyu}, DIODE~\cite{diode_dataset}, and ScanNet~\cite{scannet}. As shown in Table~\ref{tab:normal_zero_shot}, our method achieves consistently strong performance across these benchmarks. Compared with purely monocular normal estimators, our method benefits from explicit stereo geometry, which provides a more reliable starting point for refinement. In particular, the disparity-derived normal initialization supplies structured guidance before diffusion refinement, while the aligned right-view condition provides additional cross-view evidence. Together, these cues enable the model to recover sharper boundaries, more faithful local structures, and more complete surface geometry under unseen domains. Moreover, as illustrated in Fig.~\ref{fig:dn}, our method consistently improves normal estimation in diverse real indoor scenes, including challenging cases with cluttered layouts and reflective materials.

\begin{table*}[t]
\centering
\caption{Ablation study of different disparity-normal estimation designs on the IRS~\cite{irs} dataset. Variants correspond to the architectures illustrated in Fig.~\ref{fig:compare_arch}. All results are evaluated under the zero-shot setting without fine-tuning.}
\label{tab:arch_ablation}

\setlength{\tabcolsep}{5pt}
\renewcommand{\arraystretch}{1.12}
\arrayrulecolor{black}
\setlength{\arrayrulewidth}{0.8pt}

\begin{tabular}{l|cc|ccccc}
\toprule

\multirow{2}{*}{\centering Method}
& \multicolumn{2}{c|}{Disparity Error}
& \multicolumn{5}{c}{Normal Error} \\
\hhline{~|--|-----}

& EPE$\downarrow$ & D1$\downarrow$
& Mean$\downarrow$ & Median$\downarrow$ & $<11.25^\circ\uparrow$ & $<22.5^\circ\uparrow$ & $<30^\circ\uparrow$ \\

\specialrule{0.5pt}{0pt}{0pt}
\specialrule{0.5pt}{1pt}{0pt}

Baseline (LightStereo~\cite{guo2025lightstereo})
& 2.4440 & 14.35\%
& 47.04$^\circ$ & 35.45$^\circ$ & 11.13\% & 30.46\% & 42.42\% \\

+ Sequential with Feed-forward network~\cite{dpt}
& 1.5718 & 9.33\%
& 21.639$^\circ$ & 13.25$^\circ$ & 44.04\% & 67.70\% & 75.74\% \\


+ Parallel with Feed-forward network~\cite{dsine}
& \second{1.3000} & \second{7.19\%}
& \second{20.239$^\circ$} & \first{12.55$^\circ$} & \first{46.15\%} & \second{69.48\%} & \second{77.70\%} \\
\rowcolor{cyan!10}
+ Parallel with Diffusion based network~\cite{stable_diffusion}
& \first{1.1091} & \first{5.81\%}
& \first{17.56$^\circ$} & \second{13.45$^\circ$} & \second{43.80\%} & \first{70.00\%} & \first{80.03\%} \\

\bottomrule
\end{tabular}
\end{table*}

\subsection{Ablation study}
Unless otherwise specified, all ablation variants are trained with the same training set and the same number of training iterations, and are evaluated in the zero-shot regime, with only the target component changed for fair comparison.

\noindent\textbf{Ablation on Feed-forward and Diffusion-based Branches.} We further investigate how different normal-estimation designs affect the unified framework, as illustrated in Fig.~\ref{fig:compare_arch}. We compare three representative strategies: (a) a sequential design with feed-forward networks; (b) a parallel design with feed-forward networks; and (c) a parallel design that combines a diffusion-based network with a feed-forward network. All variants share the same stereo backbone for fair comparison. As shown in Table~\ref{tab:arch_ablation}, all three designs improve both disparity and normal estimation over the baseline, confirming the benefit of normal-aware modeling in the unified framework. Among them, the design with the diffusion-based network achieves the best overall performance, suggesting that its stronger priors are beneficial to both disparity and normal estimation.

\begin{table}[t]
\centering
\small
\caption{Ablation on the effect of Warp to left-view conditions for surface normal estimation on 3d Ken-Burns~\cite{3dken} and IRS~\cite{irs} datasets. We compare variants with left view, with right view, and with warp to left-view under the same setting.}

\label{tab:right_ablation}

\setlength{\tabcolsep}{3pt}

\begin{tabular}{l|ccccc}
\toprule
\multirow{2}{*}{Method} & \multicolumn{5}{c}{Normal Error} \\
\hhline{~-----}
& Mean$\downarrow$ & Median$\downarrow$ & $<11.25^\circ\uparrow$ & $<22.5^\circ\uparrow$ & $<30^\circ\uparrow$ \\
\specialrule{0.5pt}{0pt}{0pt}
\specialrule{0.5pt}{1pt}{1pt}

\multicolumn{6}{c}{3d Ken-Burns~\cite{3dken}} \\
\midrule
StableNormal~\cite{ye2024stablenormal} & \second{19.68$^\circ$} & \first{11.02$^\circ$} & \second{45.70\%} & \second{70.90\%} & \second{80.10\%} \\
Left view & 19.74$^\circ$ & 11.21$^\circ$ & 45.12\% & 70.02\% & 79.08\% \\
Right view & 20.31$^\circ$ & 12.96$^\circ$ & 43.95\% & 68.93\% & 78.02\% \\
\rowcolor{cyan!10}
Warp to left-view & \first{19.02$^\circ$} & \second{11.56$^\circ$} & \first{46.82\%} & \first{71.96\%} & \first{81.06\%} \\
\midrule

\multicolumn{6}{c}{IRS~\cite{irs}} \\
\midrule
StableNormal~\cite{ye2024stablenormal} & \second{19.01$^\circ$} & \second{11.18$^\circ$} & \second{47.02\%} & 74.06\% & 79.42\% \\
Left view & 19.08$^\circ$ & 11.31$^\circ$ & 46.71\% & \second{74.98\%} & \second{80.21\%} \\
Right view & 19.57$^\circ$ & 12.06$^\circ$ & 45.83\% & 73.84\% & 79.03\% \\
\rowcolor{cyan!10}
Warp to left-view & \first{18.63$^\circ$} & \first{10.42$^\circ$} & \first{48.94\%} & \first{75.97\%} & \first{81.18\%} \\
\bottomrule
\end{tabular}

\end{table}

\noindent\textbf{Ablation on Warp to Left-View Condition.} We analyze different right-view conditions for normal estimation on 3D Ken Burns~\cite{3dken} and IRS~\cite{irs}. As shown in Table~\ref{tab:right_ablation}, directly using the original right view does not improve over the left-view condition, mainly because the two views are spatially misaligned. In contrast, warping the right image into the left-view coordinate system achieves the best overall performance on both benchmarks. The aligned condition preserves complementary cross-view information while making it easier for the denoising U-Net to exploit, leading to sharper boundaries and more complete normal predictions.

\noindent\textbf{Ablation on Disparity to Normal Initialization.}
We evaluate the disparity to normal prior,
denoted as D2N, on the IRS~\cite{irs} dataset. As shown in
Table~\ref{tab:d2n_ablation}, removing D2N degrades both disparity
and surface normal estimation. The disparity-derived normal provides
geometry-aware guidance for the diffusion-based normal estimator.
Since it captures the scene layout, local surface orientation, and
coarse geometric discontinuities, the diffusion model can focus on
correcting local artifacts and recovering fine-grained geometric
structures, rather than inferring the surface geometry solely from
the noisy latent. Because the D2N pathway is differentiable, normal
supervision can propagate back to the stereo branch during
optimization. This transfers the structural priors learned by the
diffusion model to disparity estimation and encourages the predicted
disparity to remain geometrically consistent with the refined surface
normals, improving disparity estimation.
\begin{table}[t]
\centering
\small
\caption{
Ablation study of disparity to normal initialization on
IRS~\cite{irs}. We compare variants with and without D2N.
}
\label{tab:d2n_ablation}

\setlength{\tabcolsep}{2.5pt}
\resizebox{\columnwidth}{!}{
\begin{tabular}{l|cc|ccccc}
\toprule
\multirow{2}{*}{Method}
& \multicolumn{2}{c|}{Disparity Error}
& \multicolumn{5}{c}{Normal Error} \\
\cmidrule(lr){2-3}
\cmidrule(lr){4-8}
& EPE$\downarrow$
& D1$\downarrow$
& Mean$\downarrow$
& Median$\downarrow$
& $<11.25^\circ\uparrow$
& $<22.5^\circ\uparrow$
& $<30^\circ\uparrow$ \\
\hhline{========}

w/o D2N
& \second{1.18}
& \second{6.07\%}
& \second{19.02$^\circ$}
& \second{15.27$^\circ$}
& \second{40.19\%}
& \second{66.37\%}
& \second{76.19\%} \\

\rowcolor{cyan!10}
w/ D2N
& \first{1.11}
& \first{5.81\%}
& \first{17.56$^\circ$}
& \first{13.45$^\circ$}
& \first{43.80\%}
& \first{70.00\%}
& \first{80.03\%} \\

\bottomrule
\end{tabular}
}
\end{table}

%% file: MM26/5conclusion.tex
\section{Conclusion}

In this paper, we presented GeoStereo, a unified stereo geometry estimation framework for disparity and surface normal prediction. Our method combines a feed-forward stereo matching pipeline with a diffusion-based normal estimation branch through feature representations, disparity to normal initialization, and aligned cross-view conditioning. This design preserves task-suited modeling paradigms for the two tasks while enabling effective interaction between them. In particular, the diffusion branch introduces strong diffusion priors that enhance disparity estimation in challenging regions, while the disparity-derived initialization provides reliable geometric guidance for normal prediction, leading to more stable and consistent results for both tasks. Extensive experiments demonstrate that GeoStereo achieves strong zero-shot generalization across diverse synthetic and real-world benchmarks, and performs reliably in a wide range of challenging scenarios. In particular, in indoor scenes, our method produces more accurate geometry estimation and more complete reconstructions with clearer structures and boundaries. Despite these promising results, the framework still has room for improvement in efficiency, especially due to the additional cost introduced by the diffusion-based normal refinement branch. In future work, we plan to further improve the efficiency of the unified model and extend it to intrinsic geometric tasks, such as depth, albedo, and material-related scene properties.

\section{Acknowledgements}
This study is supported by the Beijing Academy of
Artificial Intelligence (BAAI), under its research funding
programs.